\pdfoutput=1
\documentclass{article}

\usepackage{times}
\usepackage{graphicx} % more modern
\usepackage{subfigure} 

\usepackage{natbib}

\usepackage{algorithm}
\usepackage{algorithmic}

\usepackage{amssymb}
\usepackage{amsmath}

\usepackage{hyperref}

\usepackage[accepted]{icml2016}

\icmltitlerunning{Attentive Pooling Networks}

\begin{document} 

\twocolumn[
\icmltitle{Attentive Pooling Networks}

% It is OKAY to include author information, even for blind
% submissions: the style file will automatically remove it for you
% unless you've provided the [accepted] option to the icml2016
% package.
\icmlauthor{Cicero dos Santos}{cicerons@us.ibm.com}
\icmlauthor{Ming Tan}{mingtan@us.ibm.com}
\icmlauthor{Bing Xiang}{bingxia@us.ibm.com}
\icmlauthor{Bowen Zhou}{zhou@us.ibm.com}
\icmladdress{IBM Watson,
            T.J. Watson Research Center, NY, USA}
%\icmladdress{Their Fantastic Institute,
%           27182 Exp St., Toronto, ON M6H 2T1 CANADA}

% You may provide any keywords that you 
% find helpful for describing your paper; these are used to populate 
% the "keywords" metadata in the PDF but will not be shown in the document
%\icmlkeywords{boring formatting information, machine learning, ICML}

\vskip 0.3in
]

\begin{abstract} 
In this work,
we propose \emph{Attentive Pooling} (AP),
a two-way attention mechanism for discriminative model training.
In the context of pair-wise ranking or classification with neural networks,
AP enables the pooling layer to be aware of the current input pair, 
in a way that information from the two input items can directly influence the computation of each other's representations.
Along with such representations of the paired inputs, AP jointly learns a similarity measure over projected segments (e.g. trigrams) of the pair, and subsequently, derives the corresponding attention vector for each input to guide the pooling. 
Our two-way attention mechanism is a general framework independent of the underlying representation learning,
and it has been applied to both convolutional neural networks (CNNs) and recurrent neural networks (RNNs) 
in our studies.
The empirical results, from three very different benchmark tasks of question answering/answer selection, 
demonstrate that our proposed models outperform a variety of strong baselines and achieve state-of-the-art performance in all the benchmarks.

\end{abstract} 

\section{Introduction}
\label{introduction}
Neural networks (NN) with attention mechanisms have recently proven to be successful at different computer vision (CV) and natural language processing (NLP) tasks such as image captioning \cite{xu:icml2015},
machine translation \cite{bahdanau2015:ICLR} and 
factoid question answering \cite{moritz:NIPS2015}.
%Despite the success of neural networks,
%especially convolutional neural networks (CNNs),
%for pair-wise ranking/classification tasks \cite{weston2014,Shen@CIKM2014},
However, most recent work on neural attention models have focused on one-way attention mechanisms based on recurrent neural networks designed for generation tasks.

Another important family of machine learning tasks are centered around pair-wise ranking or classification, 
which have a broad set of applications, including but not limited to, question answering, entailment, 
paraphrasing and any other pair-wise matching problems.
The current state-of-the-art models usually include NN-based representation for the input pair, followed by a discriminative ranking or classification models. For example, a convolution (or a RNN) and a max-pooling is used to independently construct distributed vector representations of the input pair, followed by a large-margin training \cite{hu2014,weston2014,Shen@CIKM2014,dos2015learning}.

The key contribution of this work is that we propose \emph{Attentive Pooling} (AP),
a two-way attention mechanism, that significantly improves such discriminative models' performance on 
pair-wise ranking or classification, by enabling a joint learning of the representations of both inputs 
as well as their similarity measurement.

Specifically, AP enables the pooling layer to be aware of the current input pair, 
in a way that information from the two input items can directly influence the computation of each other's representations.
The main idea in AP consists of learning a similarity measure over projected segments (e.g. trigrams) of the two items in the input pair,
and using the similarity scores between the segments to compute attention vectors in both directions. 
Next,
the attention vectors are used to perform pooling.

There are a few key benefits of our model.
\begin{itemize}
\item Thanks to the two-way attention, our model projects the paired inputs, even though they may not be always semantically comparable for some applications (e.g., questions and answers in question answering), into a common representation space that they can be compared in a more plausible way.
\item Our model is effective in matching pairs of inputs with significant length variations.
\item The two-way attention mechanism is independent of the underlying representation learning. For example, AP can be applied to both CNNs and RNNs, which is in contrast to the one-way attention used in the generation models mostly based on recurrent nets.
\end{itemize}

%owever, 
%despite the success of convolutional neural networks (CNNs) for classification \cite{kalchbrenner:acl2014,dosSantosCOLING2014} 
%and ranking tasks in NLP \cite{weston2014,Shen@CIKM2014}, 
%most recent work on neural attention models have focused on  recurrent neural networks.

%Recent work have demonstrated that,
%in the context of semantically equivalent question retrieval,
%CNN based representations do not scale well with the size of the input text \cite{dos2015learning}.
%Even when using deeper networks,
%with two or more convolutional layers,
%usually there is not much improvement on the final task \cite{yin2015,hu2014}.

%like answer selection and paraphrasing, where the 
%Here we argue that for tasks like answer passage selection and equivalent question retrieval,
%where the main goal is to match semantically related text pairs,
%a two-way attention mechanism can help the NN to handle larger input texts.
%The rationale is that an attention mechanism can drive the NN to
%select information that is more relevant with regard to the input pair, and not to each item separately.
%can improve the informativeness of CNN based representations by

% In this work,
% we present some preliminary results on using 
% AP on top of the hidden states of a Bidirectional Long Short-Term Memory RNN (biLSTM),
% in a similar way as max-pooling is used by Tan et al. \yrcite{tan:Arxiv15}.

In this work,
we perform an extensive number of experiments on applying attentive pooling CNNs (AP-CNN) and biLSTMs (AP-biLSTM) for the answer selection task.
In this task,
given a question $q$ and an candidate answer pool $P=\{a_1, a_2, \cdots , a_p\}$ for this question, 
the goal is to search for and select the candidate answer $a \in P$ that correctly answers $q$.
We perform experiments with three publicly available benchmark datasets, which vary in data scale, complexity and length ratios between question and answers: InsuranceQA, TREC-QA and WikiQA.
For the three datasets,
AP-CNN and AP-biLSTM respectively outperform the CNN and the biLSTM that do not use attention.
Additionally,
AP-CNN achieves state-of-the-art results for the three datasets.

Our experimental results also demonstrate that attentive pooling makes the CNN more robust to large input texts.
This is an important finding,
since recent work have demonstrated that,
in the context of semantically equivalent question retrieval,
CNN based representations do not scale well with the size of the input text \cite{dos2015learning}.
Additionally,
as AP-CNN does not rely only on the final vector representation to capture interactions between the input question and answer,
it requires much less convolutional filters than the regular CNN. It means that AP-CNN-based representations are more compact,
which can help to speed up the training process.

Although we demonstrate experimental results for NLP tasks only, 
AP is a general method that can be also applied to different types of NNs that perform matching of two inputs.
Therefore,
we believe that AP can be useful for different applications, such as computer vision and bioinformatics.

This paper is organized as follows. 
In Section \ref{neural_nets}, 
we describe two NN architectures for answer selection that have been recently proposed in the literature. 
In Section \ref{ap_networks}, 
we detail the attentive pooling approach.
In Section \ref{related_work}, 
we discuss some related work. 
Sections \ref{experimental_setup} and \ref{experimental_results} detail our experimental setup and results, respectively. 
In Section \ref{conclusions} we present our final remarks.

\section{Neural Networks for Answer Selection}
\label{neural_nets}
Different neural network architectures have  been recently proposed to perform matching of semantically related text segments \cite{yu2014,hu2014,dos2015learning,wang2015,severyn2015,tan:Arxiv15}.
In this section we briefly review two NN architectures that have previously been applied to the answer selection task:
QA-CNN \cite{feng2015applying} and QA-biLSTM \cite{tan:Arxiv15}.
Given a pair ($q$, $a$) consisting of a question $q$ and a candidate answer $a$,
both networks score the pair by first computing fixed-length independent continuous vector representations $r^q$ and $r^a$,
and then computing the cosine similarity between these two vectors.

\begin{figure}[ht]
\vskip 0.2in
\begin{center}
\centerline{\includegraphics[width=\columnwidth]{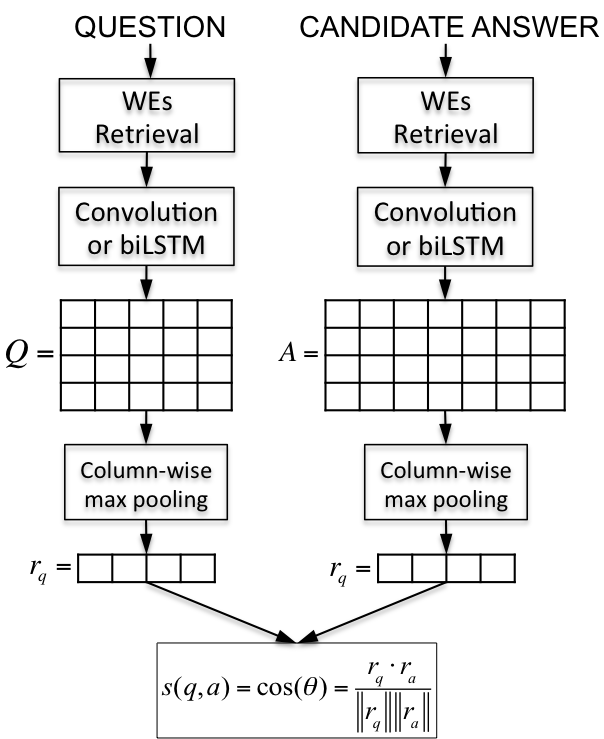}}
\caption{Joint illustration of QA-CNN and QA-biLSTM.}
\label{qacnn}
\end{center}
\vskip -0.2in
\end{figure}

In Figure \ref{qacnn}
we present a joint illustration of these two neural networks.
The first layer in both QA-CNN and QA-biLSTM transforms each input word $w$ into a fixed-size real-valued word embedding $r^{w} \in \mathbb{R}^{d}$.
Word embeddings (WEs) are encoded by column vectors in an embedding matrix $W^{0}\in\mathbb{R}^{d\times|V|}$,
where $V$ is a fixed-sized vocabulary
and $d$ is the dimention of the word embeddings.
Given the input pair ($q$, $a$),
where the question $q$ contains $M$ tokens and the candidate answer $a$ contains $L$ tokens,
the output of the first layer consists of two sequences of word embeddings $q^{emb}=\{r^{w_{1}}, ..., r^{w_{M}}\}$ and $a^{emb}=\{r^{w_{1}}, ..., r^{w_{L}}\}$.
Next,
QA-CNN and QA-biLSTM use different approaches to process these sequences.
While QA-CNN process both $q^{emb}$ and $a^{emb}$ using a convolution,
QA-biLSTM uses a Bidirectional Long Short-Term Memory RNN \cite{lstm1997} to process these sequences.

\subsection{Convolution}
Given the sequence $q^{emb}=\{r^{w_{1}}, ..., r^{w_{M}}\}$,
let us define the matrix $Z^q = [z_1,...,z_M]$ as a matrix where each column contains a vector $z_m\in\mathbb{R}^{dk}$ that is the concatenation of a sequence of $k$ word embeddings centralized in the $m$-th word of the question.
The output of the convolution with $c$ filters over the question $q$ is computed as follows:
\begin{equation} \label{conv_layer_apcnn_Q}
Q = W^{1}Z^q+ b^{1}
\end{equation}
where each column $m$ in $Q \in \mathbb{R}^{c \times M}$ contains features extracted in a context window around the $m$-th word of $q$.
The matrix $W^{1}$ and the vector $b^{1}$ are parameters to be learned.
The number of convolutional filters $c$,
and the size of the word context window $k$
are hyper-parameters to be chosen by the user.

In a similar manner,
and using the same NN parameters $W^{1}$ and $b^{1}$,
we compute $A \in \mathbb{R}^{c \times L}$, 
the output of the convolution over the candidate answer $a$.
\begin{equation} \label{conv_layer_apcnn_A}
A = W^{1}Z^a+ b^{1}
\end{equation}

\subsection{Bidirectional LSTM (biLSTM)}
Our LSTM implementation is similar to the one in \cite{graves2013} with minor modification. Given the sequence $q^{emb}=\{r^{w_{1}}, ..., r^{w_{M}}\}$, 
the hidden vector $\mathbf{h}(t)$ (with size $H$) at the time step $t$ is updated as follows:  

\begin{eqnarray}
i_{t} & = & \sigma(\mathbf{W}_{i}r^{w_{t}}+\mathbf{U}_{i}\mathbf{h}(t-1)+\mathbf{b}_{i})\\
f_{t} & = & \sigma(\mathbf{W}_{f}r^{w_{t}}+\mathbf{U}_{f}\mathbf{h}(t-1)+\mathbf{b}_{f})\\
o_{t} & = & \sigma(\mathbf{W}_{o}r^{w_{t}}+\mathbf{U}_{o}\mathbf{h}(t-1)+\mathbf{b}_{o})\\
\tilde{C}_{t} & = & \tanh(\mathbf{W}_{m}r^{w_{t}}+\mathbf{U}_{m}\mathbf{h}(t-1)+\mathbf{b}_{m})\\
C_{t} & = & i_{t}*\tilde{C}_{t}+f_{t}*C_{t-1}\\
\mathbf{h}_{t} & = & o_{t}*\tanh(C_{t})
\end{eqnarray}

In the LSTM architecture, there are three gates
(input $i$, forget $f$ and output $o$), and a cell memory vector $c$. 
$\sigma$ is the $sigmoid$ function. The input gate can determine how incoming vectors $r^{w_{t}}$ alter the state of the memory cell. The output gate can allow the memory cell to have an effect on the outputs. Finally, the forget gate allows the cell to remember or forget its previous state.
$\mathbf{W} \in R^{H \times d}$, $\mathbf{U} \in R^{H \times H}$ and $\mathbf{b} \in R^{H \times 1}$ are the network parameters. 

Single direction LSTMs suffer a weakness of not utilizing the contextual information from the future tokens. Bidirectional LSTM utilizes both the previous and future context by processing the sequence on two directions, and generate two independent sequences of LSTM output vectors. One processes the input sequence in the forward direction, while
the other processes the input in the reverse direction.
The output at each time step is the concatenation of the two output vectors from both directions, ie. $h_t$ = $\overrightarrow{h_{t}} \parallel \overleftarrow{h_{t}}$.
We define $c=2 \times H$ for the notation consistency with the previous subsection.
After computing the hidden state $h_t$ for each time step $t$, we generate the matrices  $Q \in \mathbb{R}^{c \times M}$ and  $A \in \mathbb{R}^{c \times L}$,
where the $j$-th column in $Q$ ($A$) corresponds to $j$-th hidden state $h_j$ that is computed by the biLSTM when processing $q$ ($a$). 
The same network parameters are used to process both questions and candidate answers. 

\subsection{Scoring and Training Procedure}
Given the matrices $Q$ and $A$, 
we compute the vector representations $r^{q} \in \mathbb{R}^{c}$ and $r^{a} \in \mathbb{R}^{c}$ by applying a column-wise max-pooling over $Q$ and $A$, 
followed by a non-linearity.
Formally, 
the $j$-th elements of the vectors $r^q$ and $r^a$ are compute as follows:
\begin{equation}
[r^{q}]_j =\tanh\left(\max_{1 < m < M}\left[Q_{j,m}\right]\right)
\end{equation}
\begin{equation}
[r^{a}]_j =\tanh\left(\max_{1 < l < L}\left[A_{j,l}\right]\right)
\end{equation}

The last layer in QA-CNN and QA-biLSTM scores the input pair ($q$,$a$) by  computing the cosine similarity between the two representations:
\begin{equation} \label{cosine_sim}
s(q,a)=\frac{r^q.r^a}{\|r^q\| \|r^a\|}
\end{equation}

Both networks are trained by minimizing a pairwise ranking loss function over the training set $D$.
The input in each round is two pairs ($q$, $a^{+}$) and ($q$, $a^{-}$),
where $a^{+}$ is a ground truth answer for $q$,
and $a^{-}$ is an incorrect answer.
As in \cite{weston2014,hu2014}, 
we define the training objective as a hinge loss:
\begin{equation}
L = \max \{ 0, m - s_{\theta}(q, a^+) + s_{\theta}(q, a^-) \}
\end{equation}
where $m$ is constant margin,
$s_{\theta}(q, a^+)$ and $ s_{\theta}(q, a^-)$
are scores generated by the network with parameter set $\theta$.
During training,
for each question we randomly sample 50 negative answers from the entire answer set,
but only use the one with the highest score to update the model.

We use stochastic gradient descent (SGD) to minimize the
loss function with respect to $\theta$. 
The backpropagation algorithm is used to compute the gradients of the network.
\section{Attentive Pooling Networks for Answer Selection}
\label{ap_networks}
% In QA-CNN, 
% $r^q$ and $r^a$ are independently created as the result of a max-pooling on top of a convolution over $q$ and $a$, 
% respectively.
\emph{Attentive pooling} is an approach that enables the pooling layer to be aware of the current input pair, 
in a way that information from the question $q$ can directly influence the computation of the answer representation $r^a$, 
and vice versa.
The main idea consists of learning a similarity measure over the projected segments in the input pairs,
and uses the similarity scores between the segments to compute attention vectors.
When AP is applied to CNN,
which we call AP-CNN,
the network learns the similarity measure over the convolved input sequences.
When AP is applied to biLSTM,
which we call AP-biLSTM,
the network learns the similarity measure over the hidden states produced by the biLSTM when processing the two input sequences.
We use a similarity measure that has a bilinear form but followed by a non-linearity.

\begin{figure*}[ht]
\vskip 0.2in
\begin{center}
\centerline{\includegraphics[width=0.52\textwidth]{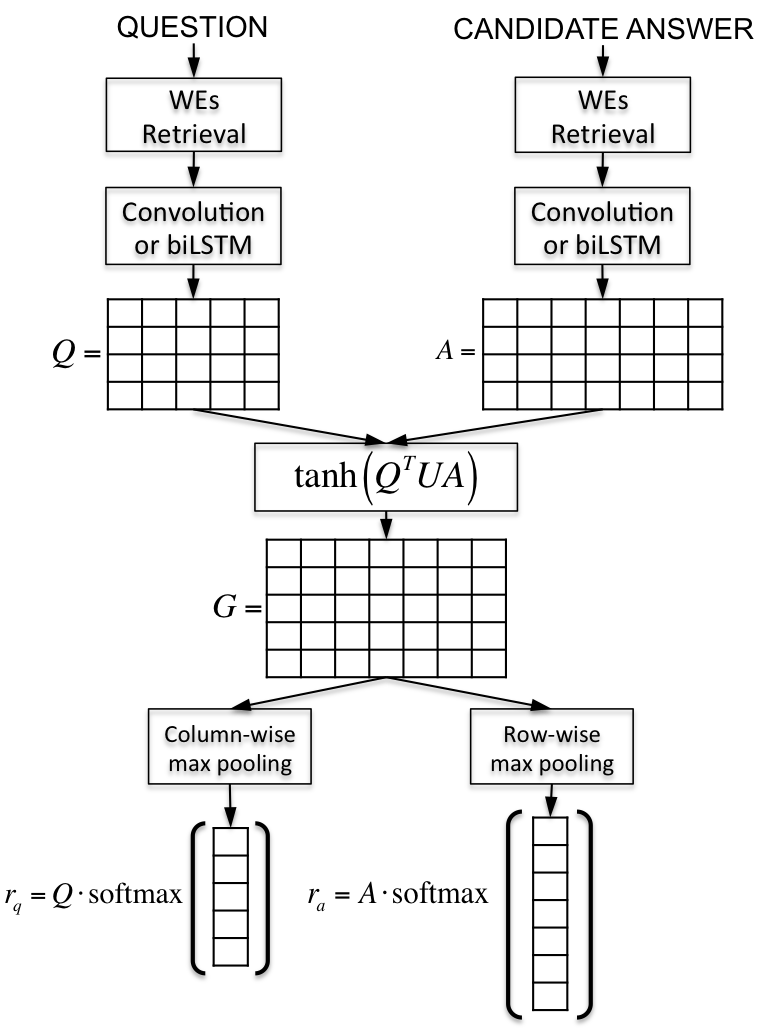}}
\caption{Attentive Pooling Networks for Answer Selection.}
\label{apcnn}
\end{center}
\vskip -0.2in
\end{figure*}

In Fig. \ref{apcnn}, 
we illustrate the application of AP over the output of the convolution or the biLSTM to construct the representations $r^q$ and $r^a$.
Consider the input pair ($q$, $a$) where the question has size $M$ and the answer has size $L$\footnote{In Fig. \ref{apcnn}, $q$ has a size of five and $a$ has a size of seven.}.
After we compute the matrices $Q \in \mathbb{R}^{c \times M}$ and  $A \in \mathbb{R}^{c \times L}$,
either by convolution or biLSTM,
we compute the matrix $G \in \mathbb{R}^{M \times L}$  as follows:
\begin{equation} \label{computing_g}
G = \tanh \left( Q^{T}UA \right)
\end{equation}
where $U \in \mathbb{R}^{c \times c}$ is a matrix of parameters to be learned by the NN.
When the convolution is used to compute $Q$ and $A$,
the matrix $G$ contains the scores of a \emph{soft alignment} between the convolved $k$-size context windows of $q$ and $a$.
When the biLSTM is used to compute $Q$ and $A$,
the matrix $G$ contains the scores of a \emph{soft alignment} between the hidden vectors of each token in $q$ and $a$.

Next, 
we apply column-wise and row-wise max-poolings over $G$ to generate the vectors $g^q \in \mathbb{R}^{M}$ and $g^a \in \mathbb{R}^{L}$, 
respectively.
Formally, 
the $j$-th elements of the vectors $g^q$ and $g^a$ are computed as follows:
\begin{equation}
[g^{q}]_j =\max_{1 < m < M}\left[G_{j,m}\right]
\end{equation}
\begin{equation}
[g^{a}]_j =\max_{1 < l < L}\left[G_{l,j}\right]
\end{equation}
We can interpret each element $j$ of the vector $g^a$ as an \emph{importance score} for the context around the $j$-th word in the candidate answer $a$ with regard to the question $q$.
Likewise,
each element $j$ of the vector $g^q$ can be interpreted as the importance score for the context around the $j$-th word in the question $q$ with regard to the candidate answer $a$.

Next, 
we apply the softmax function to the vectors $g^q$ and $g^a$ to create attention vectors $\sigma^{q}$ and $\sigma^{a}$.
For instance, 
the $j$-th element of the vector $\sigma^{q}$ is computed as follows:
\begin{equation}
[\sigma^{q}]_j = \dfrac{e^{[g^{q}]_j}}{\displaystyle \sum_{1 < l < M}{e^{[g^{q}]_l}}}
\end{equation}

Finally,
the representations $r^q$ and $r^a$ are computed as the dot product between the attention vectors $\sigma^{q}$ and $\sigma^{a}$ and the output of the convolution (or biLSTM) over $q$ and $a$, respectively:
\begin{equation} \label{sigmaq}
r^q=Q\sigma^{q}
\end{equation}
\begin{equation} \label{sigmaa}
r^a=A\sigma^{a}
\end{equation}

Like in QA-CNN and QA-biLSTM,
the final score is also computed using the cosine similarity between $r^q$ and $r^a$.
We use SGD to train AP-CNN and AP-biLSTM by minimizing the same pairwise loss function used in QA-CNN and QA-biLSTM.

\section{Related Work}
\label{related_work}
Traditional work on answer selection have normally used feature engineering, 
linguistic tools, 
or external
resources \cite{yih2013,wangmengqiu2010,wangmengqiu2007}. Recently,
deep learning (DL) approaches have been  exploited for this task and achieved significant out-performance compared to traditional non-DL methods. For example, in \cite{yu2014,feng2015applying,severyn2015}, the authors generate the representations of questions and answers separately, and score a QA pair using a similarity metric on top of these representations. In \citet{wang2015},
first a joint feature vectors is learned from a joint long short-term memory (LSTM) model connecting questions and answers, 
and then the task is converted into a learning-to-rank problem. 

At the same time, attention-based systems have shown very promising results on a variety of NLP tasks, such as machine translation \cite{bahdanau2015:ICLR,sutskever2014}, caption generation \cite{xu:icml2015} and factoid question answering \cite{moritz:NIPS2015}. 
Such models learn to focus their “attention” to specific parts of their input. 

Some recently-proposed approaches introduce  attention mechanisms in the answer selection task. Tan et al. \yrcite{tan:Arxiv15} developed an attentive reader based on bidirectional long short-term memory, which emphasizes certain part of the answer according to the question embedding. Unlike \cite{tan:Arxiv15}, in which attention is imposed only on answer embedding generation, AP-CNN and AP-biLSTM consider the interdependence between questions and answers.

In the context of two-way attention,
two very recent work are related to ours.
Rockt{\"{a}}schel et al. \yrcite{Rocktaschel:reasoning15},
propose a two-way attention method that is inspired by bidirectional LSTMs that read a sequence and its reverse for improved encoding.
Their approach,
which is designed for RNNs only,
differs in many aspects from the approach described in this work,
which can be easily applied for CNNs and RNNs.
Yin et al. \yrcite{yin2015} present a two-way attention mechanism that is tailored to CNNs.
Some of the main differences between their approach and this work are: (1) they use a simple Euclidean distance to compute the interdependence between the two input texts, while in this work we apply similarity metric learning, which has the potential to learn better ways to measure the interaction between segments of the input items; (2) the models in \cite{yin2015} compute the attention vector using sum-pooling over the alignment matrix and use the convolutional outputs updated by the attention as the input for another level of convolutional layer. In this work we use max-pooling over the alignment matrix plus softmax, in order to explicitly create an attention vector that is used to perform the pooling. Experimental results show that such difference yields substantial improvement of performance on WikiQA dataset. 
\section{Experimental Setup}
\label{experimental_setup}
%In this section we detail our experimental setup.

\subsection{Datasets}
We apply AP-CNN, 
AP-biLSTM,
QA-CNN and
QA-biLSTM to three different answer selection datasets:
InsuranceQA,
TREC-QA
and WikiQA.
These datasets contain text of different domains and have different caracteristics.
Table \ref{as:datasets} presents some statistics about the datasets, including the number of questions in each set, average length of questions (M) and answers (L), average number of candidate answers in the dev/test sets and the average ratio between the lengths of questions and their ground-truth answers. 

\begin{table*}[ht]
\caption{Answer Selection Datasets.}
\label{as:datasets}
\vskip 0.15in
\begin{center}
\begin{small}
\begin{sc}
\begin{tabular}{lrrrcccc}
\hline
\abovespace\belowspace
Dataset  & Train & Dev  & Test  & Avg. M  & Avg. L & Avg. \# Cand. Ans. & Avg. L/M \\
\hline
\abovespace
InsuranceQA & 12887 & 1000 & 1800x2 & 7 & 95 & 500 & 13.8 \\ 
TREC-QA      & 1162 & 65 & 68  & 8 & 28  & 38 & 4.2 \\ 
WikiQA       & 873 & 126 & 243 & 6 & 25 & 9 & 5.0 \\
\hline
\end{tabular}
\end{sc}
\end{small}
\end{center}
\vskip -0.1in
\end{table*}

InsuranceQA\footnote{git clone https://github.com/shuzi/insuranceQA.git} is a recently released large-scale non-factoid QA dataset from the insurance domain. This dataset provides a training set, a validation set, and two test sets. We do not see obvious categorical differentiation between questions of the two test sets. For each question in dev/test sets, there is a set of 500 candidate answers, which include the ground-truth answers and randomly selected negative answers. More details can be found in \cite{feng2015applying}. 

TREC-QA\footnote{The data is obtained from \cite{yao2013} \url{http://cs.jhu.edu/~xuchen/packages/jacana-qa-naacl2013-data-results.tar.bz2}}
was created by Wang et al. \yrcite{wangmengqiu2007} based on Text REtrieval Conference
(TREC) QA track (8-13) data. We follow the exact approach of train/dev/test questions selection in
\cite{wang2015}, in which all questions with only positive or negative answers are removed.
Finally, we have 1162 training questions, 65 development questions and 68 test questions.

WikiQA\footnote{The data is obtained from \cite{yang2015}} is an open domain question-answering
dataset. We use the subtask that assumes that there is at least one correct answer for a question. The
corresponding dataset consists of 20,360 question/candidate pairs in train, 
1,130 pairs in dev and 2,352
pairs in test.  
We adopt the standard setup of only considering questions that have correct answers
for evaluation. 
%Following \cite{yang2015}, we truncate answers to 40 tokens.

\subsection{Word Embeddings}
In order to fairly compare our results with the ones in previous work,
we use two different sets of pre-trained word embeddings.
For the InsuranceQA dataset,
we use the 100-dimensional vectors that were trained by Feng et al. \yrcite{feng2015applying} using word2vec \cite{word2vec2013}.
Following Wang \& Nyberg \yrcite{wang2015},
Tan et al. \yrcite{tan:Arxiv15} and Yin et al.\yrcite{yin2015},
for the TREC-QA and the WikiQA datasets we use the 300-dimensional vectors that were trained using word2vec and are publicly available on the website of this tool\footnote{https://code.google.com/p/word2vec/}.

\subsection{Neural Networks Setup}
In Table \ref{tab:nn_hyperparams}, 
we show the selected hyperparameter values, 
which were tuned using the validation sets.
We try to use as much as possible the same hyperparameters for all the three datasets.
The size of the word embeddings is different due to the different pre-trained versions that we used for InsuranceQA and the other two datasets.
We use a context window of size 3 for InsuranceQA,
while we set this parameter to 4 for TREC-QA and WikiQA.
Using the selected hyperparameters,
the best results are normally achieved using between 15 and 25 training epochs.
For AP-CNN,
AP-biLSTM and QA-LSTM,
we also use a learning rate schedule that decreases the learning rate $\lambda$ according to the training epoch $t$.
Following \citet{santos2014},
we set the learning rate for epoch $t$, 
$\lambda_t$, 
using the equation:
$\lambda_t = \dfrac{\lambda}{t}$.

\begin{table*}[ht!]
\caption{Neural Network Hyper-Parameters}
\label{tab:nn_hyperparams}
\vskip 0.15in
\begin{center}
\begin{small}
\begin{sc}
\begin{tabular}{llrrrr}
\hline
\abovespace\belowspace
\bf Hyp. & \bf Hyperpar. Name  & \bf AP-CNN & \bf QA-CNN & \bf AP-biLSTM & \bf QA-biLSTM \\
\hline
$d$          & Word Emb. size        & 100/300 & 100/300 & 100/300 & 100/300 \\
$c$          & Conv. Filters / Hid.Vec. Size         & 400     & 4000    & 141x2 & 141x2 \\ 
$k$          & Context Window size   & 3/4     & 2       & 1 & 1 \\ 
$mbs$        & Minibatch size        & 20      & 1       & 20 & 20 \\
$m$          & Loss margin           & 0.5     & 0.009   & 0.2 & 0.1 \\
$\lambda$    & Init. Learning Rate & 1.1       & 0.05    & 1.1 & 1.1 \\
\hline
\end{tabular}
\end{sc}
\end{small}
\end{center}
\vskip -0.1in
\end{table*}

In our experiments, 
the four NN architectures QA-CNN, 
AP-CNN, 
QA-biLSTM 
and AP-biLSTM are implemented using Theano \cite{bergstra:scipy2010}.

\section{Experimental Results}
\label{experimental_results}

\subsection{InsuranceQA}
In Table \ref{tab:insuranceqa},
we present the experimental results of the four NNs for the InsuranceQA dataset.
The results are in terms of accuracy,
which is equivalent to precision at top one.
On the bottom part of this table,
we can see that AP-CNN outperforms QA-CNN by a large margin in both test sets,
as well as in the dev set.
AP-biLSTM also outperforms the QA-biLSTM in all the three sets.
AP-CNN and AP-biLSTM have similar performance.

On the top part of Table \ref{tab:insuranceqa} we present the results of two state-of-the-art systems for this dataset.
In \cite{feng2015applying},
the authors present a CNN architecture that is similar to QA-CNN,
but that uses a different similarity metric instead of cosine similarity.
In \cite{tan:Arxiv15},
the authors use a biLTSM architecture that employs unidirectional attention.
Both AP-CNN and AP-biLSTM outperform the state-of-the-art systems.

\begin{table}[ht!]
\caption{Accuracy of different systems for InsuranceQA}
\label{tab:insuranceqa}
\vskip 0.15in
\begin{center}
\begin{small}
\begin{sc}
\begin{tabular}{lrrr}
\hline
\abovespace\belowspace
\bf System & \bf Dev  & \bf Test1 & \bf Test2 \\
\hline
%Bag-of-word & 31.9 & 32.1 & 32.2 \\
%MB-IR & 52.7 & 55.1 & 50.8 \\
\cite{feng2015applying} & 65.4 & 65.3  & 61.0 \\ 
\cite{tan:Arxiv15}  & 68.4 & 68.1 & 62.2 \\
\hline
QA-CNN    & 61.6 & 60.2 & 56.1\\
QA-biLSTM & 66.6 & 66.6 & 63.7 \\
AP-CNN    & \bf 68.8  &  69.8  & 66.3 \\
AP-biLSTM & 68.4 &  \bf 71.7 &  \bf 66.4\\
\hline
\end{tabular}
\end{sc}
\end{small}
\end{center}
\vskip -0.1in
\end{table}
%\multirow{2}{*}{65.4} 

One important characteristic of AP-CNN is that it requires less convolutional filters than QA-CNN.
For the InsuranceQA dataset,
AP-CNN uses 10x less filters (400) than  QA-CNN (4000).
Using 800 filters in AP-CNN produces very similar results as using 400.
On the other hand,
as also found in \cite{feng2015applying},
QA-CNN requires at least 2000 filters to achieve more than 60\% accuracy on InsuranceQA.
AP-CNN needs less filters because it does not rely only on the final vector representation to capture interactions between the input question and answer.
As a result,
although AP-CNN has a more complex architecture,
its training time is two times faster than QA-CNN.
Using a Tesla K20Xm,
our Theano implementation of AP-CNN takes about 16 minutes to complete one epoch (training + inference over validation set) for InsuranceQA,
which consists on processing 1.5 million text segments.

\begin{figure}[ht]
\vskip 0.2in
\begin{center}
\centerline{\includegraphics[width=\columnwidth]{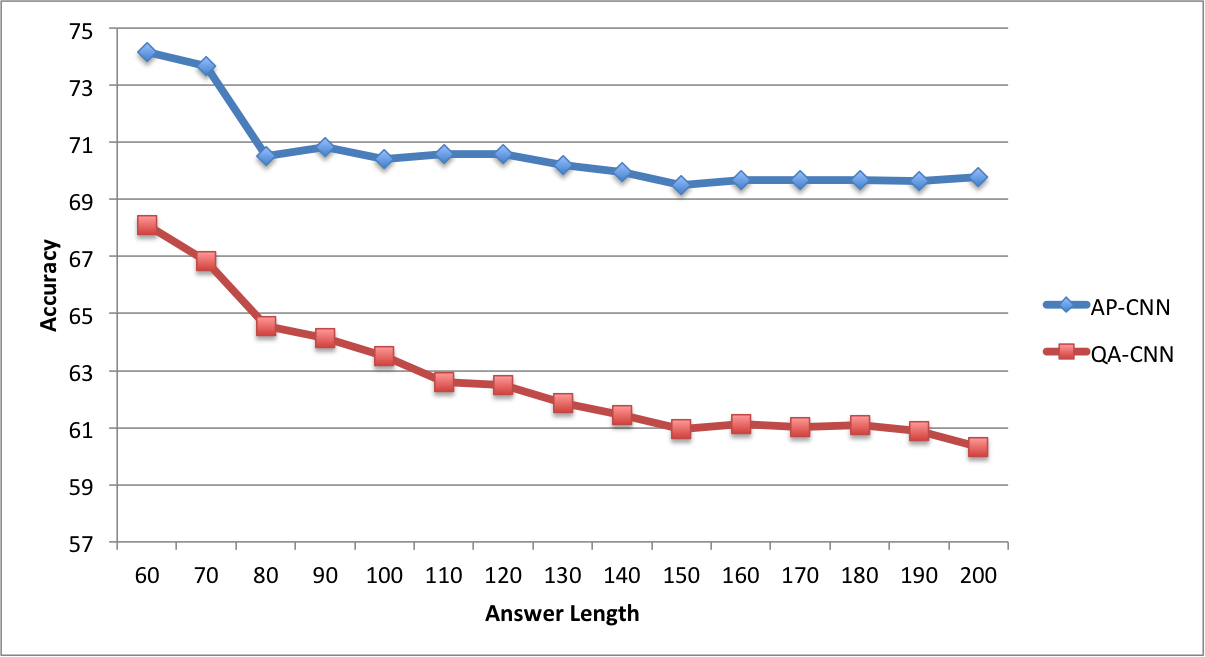}}
\caption{Aggregated  accuracy for answers
up to a certain length in the InsuranceQA Test1 set}
\label{apcnn_vs_qacnn_t1}
\end{center}
\vskip -0.2in
\end{figure}

\begin{figure}[ht]
\vskip 0.2in
\begin{center}
\centerline{\includegraphics[width=\columnwidth]{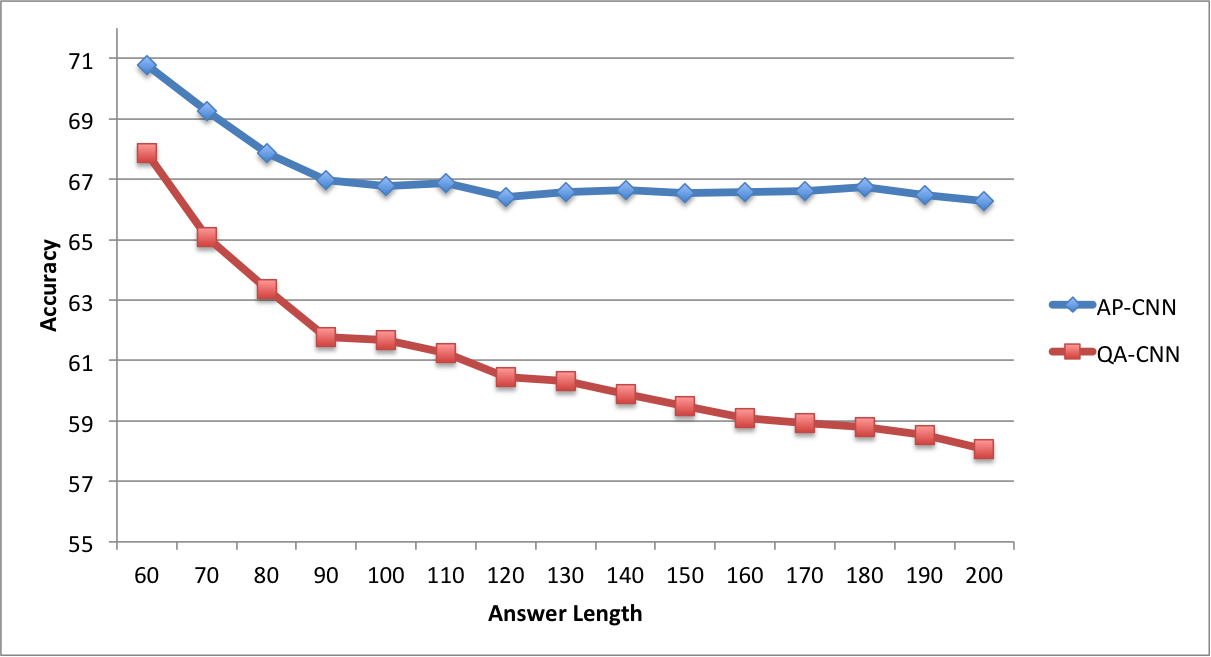}}
\caption{Aggregated accuracy for answers
up to a certain length in the InsuranceQA Test2 set}
\label{apcnn_vs_qacnn_t2}
\end{center}
\vskip -0.2in
\end{figure}

In figures \ref{apcnn_vs_qacnn_t1} and \ref{apcnn_vs_qacnn_t2},
we plot the aggregated accuracy of AP-CNN and QA-CNN for answers up to a certain length for the Test1 and Test2 sets,
respectively.
We can see in both plots that the performance of both system is better for shorter answers.
However,
while the performance of QA-CNN continues to drop as larger answers are considered,
the performance of AP-CNN seems to be stable after reaching a length of $\sim$90 tokens.
These results give support to our hypothesis that attentive pooling helps the CNN to become robust to larger input texts.

\subsection{TREC-QA}
In Table \ref{tab:trecqa},
we present the experimental results of the four NNs for the TREC-QA dataset.
The results are in terms of mean average precision (MAP) and mean reciprocal rank (MRR),
which are the metric normally used in previous work with the same dataset.
We use the official \emph{trec\_eval} scorer to compute MAP and MRR.
We can see in Table \ref{tab:trecqa} that AP-CNN outperforms QA-CNN by a large margin in both metrics.
AP-biLSTM outperforms the QA-biLSTM,
but its performance is not as good as the of AP-CNN.

On the top part of Table \ref{tab:trecqa} we present the results of three recent work that use TREC-QA as a benchmark.
In \cite{wang2015},
the authors present an LTSM architecture for answer selection. Their best result consists of a combination of LSTM and the BM25 algorithm.
In \cite{severyn2015},
the authors propose an NN architecture where the representations created by a convolutional layer are the input to similarity measure learning.
Wang \& Ittycheriah \yrcite{wang_Ittycheriah2015} propose a word-alignment-based method that is suitable for the FAQ-based QA task.
AP-CNN outperforms the state-of-the-art systems in both metrics,
MAP and MRR.

\begin{table}[ht!]
\caption{Performance of different systems for TREC-QA}
\label{tab:trecqa}
\vskip 0.15in
\begin{center}
\begin{small}
\begin{sc}
\begin{tabular}{lrr}
\hline
\abovespace\belowspace
\bf System & \bf MAP & \bf MRR \\
\hline
\citet{wang2015}             & 0.7134 & 0.7913 \\
\citet{severyn2015}          & 0.7460 & 0.8080 \\
\citet{wang_Ittycheriah2015} & 0.7460 & 0.8200 \\
\hline
QA-biLSTM  & 0.6750 & 0.7723 \\
QA-CNN     & 0.7147 & 0.8070 \\ 
AP-biLSTM  & 0.7132 & 0.8032 \\
AP-CNN     & \bf 0.7530  & \bf 0.8511 \\
\hline
\end{tabular}
\end{sc}
\end{small}
\end{center}
\vskip -0.1in
\end{table}

\subsection{WikiQA}

Table \ref{tab:wikiqa} shows the experimental results of the four NNs for the WikiQA dataset.
Like in the other two datasets,
AP-CNN outperforms QA-CNN,
and AP-biLSTM outperforms the QA-biLSTM.
The difference of performance between AP-CNN and QA-CNN is smaller than the one for the InsuranceQA dataset. We believe it is because the average size of the answers in WikiQA (25) is much smaller than in InsuranceQA (95).
It is expected that attentive pooling bring more impact to the datasets that have larger answer/question lengths.

In Table \ref{tab:wikiqa} we also present the results of two recent work that use WikiQA as a benchmark.
Yang et al. \yrcite{yang2015},
present a bigram CNN model with average pooling. 
In \cite{yin2015},
the authors propose an attention-based CNN.
In order to make a fair comparison,
in Table \ref{tab:wikiqa} we include Yin et al.'s result that use word embeddings only\footnote{Yin et al. \cite{yin2015} report 0.6921(MAP) and 0.7108(MRR) when using handcrafted features in addition to word embeddings.}.
AP-CNN outperforms these two systems in both metrics.

\begin{table}[ht!]
\caption{Performance of different systems for WikiQA}
\label{tab:wikiqa}
\vskip 0.15in
\begin{center}
\begin{small}
\begin{sc}
\begin{tabular}{lrr}
\hline
\abovespace\belowspace
\bf System & \bf MAP & \bf MRR \\
\hline
\citet{yang2015}  &  0.6520   & 0.6652 \\
\citet{yin2015}   &    0.6600   & 0.6770 \\ 
\hline
QA-biLSTM  &   0.6557    &  0.6695 \\
QA-CNN     &   0.6701    &  0.6822 \\ 
AP-biLSTM  &   0.6705    &  0.6842  \\
AP-CNN     & \bf 0.6886  & \bf 0.6957 \\

\hline
\end{tabular}
\end{sc}
\end{small}
\end{center}
\vskip -0.1in
\end{table}

\subsection{Attentive Pooling Visualization}
\begin{figure}[ht]
\vskip 0.2in
\begin{center}
\centerline{\includegraphics[width=\columnwidth]{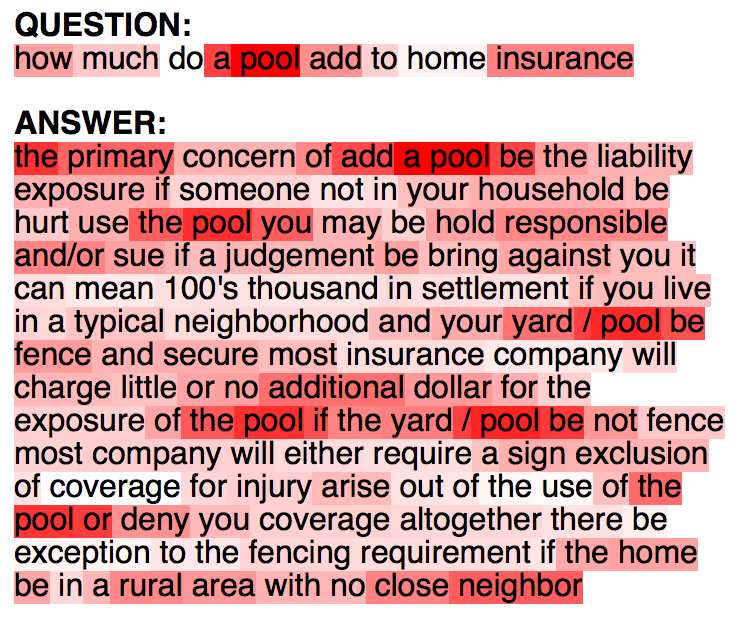}}
\caption{Attention heat map from AP-CNN for a correctly selected answer.}
\label{example1}
\end{center}
\vskip -0.2in
\end{figure}

\begin{figure}[ht]
\vskip 0.2in
\begin{center}
\centerline{\includegraphics[width=\columnwidth]{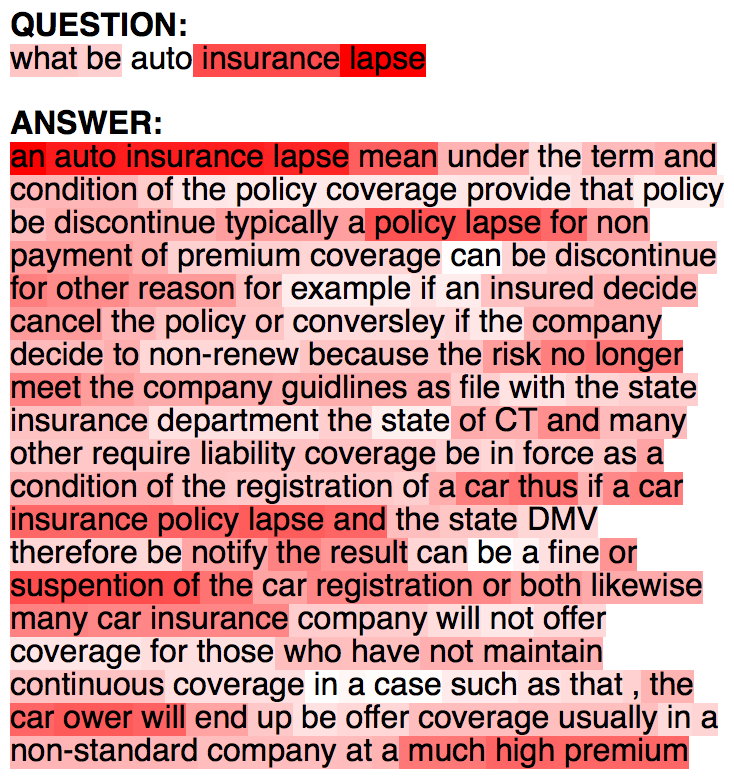}}
\caption{Attention heat map from AP-CNN for a correctly selected answer.}
\label{example2}
\end{center}
\vskip -0.2in
\end{figure}

Figures \ref{example1} and \ref{example2} depict two heat maps of two test questions from InsuranceQA that were correctly answered by AP-CNN and whose answers are more than 100 words long.
The stronger the color of a word in the question  (answer),
the larger the attention weight in $\sigma^q$ ($\sigma^a$) of the trigram centered at that word.
As we can see in the pictures,
the attentive pooling mechanism is indeed putting more focus on the segments of the answer that have some interaction with the question, and vice-verse.

% \begin{figure}[ht]
% \vskip 0.2in
% \begin{center}
% \centerline{\includegraphics[width=\columnwidth]{example2_green}}
% \caption{Attention heat map from AP-CNN for a correctly selected answer.}
% \label{example1}
% \end{center}
% \vskip -0.2in
% \end{figure}

% \begin{figure}[ht]
% \vskip 0.2in
% \begin{center}
% \centerline{\includegraphics[width=\columnwidth]{example1_green}}
% \caption{Attention heat map from AP-CNN for a correctly selected answer.}
% \label{example2}
% \end{center}
% \vskip -0.2in
% \end{figure}

\section{Conclusions}
\label{conclusions}
We present attentive pooling,
a two-way attention mechanism for discriminative model training.
The main contributions of the paper are:
(1) AP is more general than recently proposed two-way attention mechanism because: (a) it learns how to compute interactions between the items in the input pair; and (b) it can be applied to both CNNs and RNNs;
(2) we demonstrate that AP can be effectively used with CNNs and biLSTM in the context of the answer selection task, using three different benchmark datasets;
(3) our experimental results demonstrate that AP helps the CNN to cope with large input texts;
(4) we present new state-of-the-art results for InsuranceQA and TREC-QA datasets.
(5) for the WikiQA dataset
our results are the best reported so far for methods that do not use handcrafted features.

% Acknowledgements should only appear in the accepted version. 
\section*{Acknowledgements} 
The authors would like to thank Piero Molino for creating the script used to produce the text heat maps presented in this work.
 
%\textbf{Do not} include acknowledgements in the initial version of
%the paper submitted for blind review.

\bibliography{icml2016_apn}
\bibliographystyle{icml2016}

\end{document}